\documentclass[runningheads]{llncs}
\usepackage{graphicx}
\usepackage{algorithm}
\usepackage{algorithmicx}
\usepackage{algpseudocode}
\usepackage{amsmath}

\usepackage{subfig}
\begin{document}
\title{Investigating Poor Performance Regions of Black Boxes: LIME-based Exploration in Sepsis Detection}
\author{Mozhgan Salimiparsa\inst{1}\orcidID{0000-0002-0162-032X} \and
Surajsinh Parmar\inst{1}\orcidID{0009-0005-5463-0563} \and
San Lee\inst{1}\orcidID{0009-0001-6488-7988} \and
Choongmin Kim\inst{2}\orcidID{0000-0001-5812-889X} \and
Yonghwan Kim\inst{2}\orcidID{0000-0003-3022-6847} \and
Jang Yong Kim\inst{3}\orcidID{0000-0001-8437-9254}}
\authorrunning{M. Salimiparsa et al.}
\institute{SpassMed Inc., Toronto, Canada \\
\email{\{mozhgan.salimiparsa, suraj.parmar, sanlee\}@spassmed.ca}\and
Spass Inc., Seoul, Korea\\
\email{\{cmkim, kyh\}@spass.ai} \and
St. Mary's Hospital, Seoul, Korea\\
\email{vasculakim@catholic.ac.kr}}
\titlerunning{LIME-based Exploration in Sepsis Detection}% Part of RIGHT running header
\maketitle              % typeset the header of the contribution

\begin{abstract}
Interpreting machine learning models remains a challenge, hindering their adoption in clinical settings. This paper proposes leveraging Local Interpretable Model-Agnostic Explanations (LIME) to provide interpretable descriptions of black box classification models in high-stakes sepsis detection. By analyzing misclassified instances, significant features contributing to suboptimal performance are identified. The analysis reveals regions where the classifier performs poorly, allowing the calculation of error rates within these regions. This knowledge is crucial for cautious decision-making in sepsis detection and other critical applications. The proposed approach is demonstrated using the eICU dataset, effectively identifying and visualizing regions where the classifier underperforms. By enhancing interpretability, our method promotes the adoption of machine learning models in clinical practice, empowering informed decision-making and mitigating risks in critical scenarios.


\end{abstract}
\section{Introduction}
Machine learning has exhibited impressive achievements in diverse fields, including healthcare \cite{rajkomar2019machine}. The complexity of these models, however, creates challenges for their adoption in healthcare \cite{antoniadi2021current}. To address this issue, eXplainable AI (XAI) has been introduced, enabling machine learning models to provide explanations for their predictions. Model explainability is essential for gaining a deeper understanding of a model's decision-making process \cite{adadi2018peeking}. In critical domains such as sepsis detection \cite{fleuren2020machine} in the ICU, where incorrect predictions can result in fatal consequences, the reliability of these models is of utmost significance. This paper aims to tackle a specific aspect of the interpretability challenge associated with these models, specifically the identification and explanation of scenarios in which black box predictive models fail or exhibit unexpected performance.

Examining instances in which machine learning models exhibit deviations from their usual performance holds significant importance. These insights empower decision-makers to exercise caution in deploying models in situations where their predictions are prone to errors, thereby mitigating potential adverse consequences. Previous research endeavors have primarily centered on assessing the overall performance of these models through the adoption of evaluation metrics and methodologies aimed at gauging their reliability \cite{cerqueira2020evaluating,flach2019performance}. W. Duivesteijn et al. \cite{duivesteijn2014understanding} present an evaluation method that assesses the performance of a classifier, highlighting subspaces  where the classifier excels or struggles in classification tasks, however, the method's applicability is limited to binary datasets and lacks model agnosticism. L. Torgo et al. \cite{torgo2021beyond} propose approaches that aim to offer interpretable descriptions of expected performance; however, the proposed visualization may not be well-suited when dealing with a high number of features. This paper provides an analysis by focusing on the identification of specific regions where the models exhibit significant deviations from their usual performance. The identification of these regions empowers healthcare practitioners to make informed decisions by exercising caution when relying on the model. Additionally, these findings offer valuable insights that can guide the development of potential strategies aimed at improving and refining the model's overall performance \cite{roshan2021utilizing,torgo2021beyond}.

To achieve this, we propose an analytical approach that combines visual techniques to identify regions in the input space where the models' performance significantly diverges from their average performance. This visualization empowers users to grasp how various values of a particular predictor impact the models' performance.

\section{Methodology}

In this study, we adopted a modified visualization approach inspired by L. Torgo et al. \cite{torgo2021beyond} to identify the regions where a black-box model exhibits poor performance. L. Torgo et al. utilized the confusion matrix and cross-validation, and employed error distribution plots for each individual feature to demonstrate areas of inadequate model performance. However, we recognized the challenge of visual clutter arising from a large number of features. To address this limitation, we focused our analysis on identifying recurrent conditions associated with misclassifications, rather than visualizing misclassifications for each individual feature. 
To achieve this, we employed a rule extraction method, specifically LIME (Local Interpretable Model-agnostic Explanations) \cite{lime}.  LIME is a model-agnostic explainability technique that assigns importance weights to features, indicating their contribution to individual predictions. We applied LIME to misclassified data samples, allowing us to pinpoint the specific features responsible for incorrect predictions made by the classifier. This process was performed for each misclassified sample, enabling us to accumulate the features with high importance. We then intersected these features, focusing on those that consistently appeared as contributing factors to misclassifications. This allowed us to discern regions or intervals in which the classifier demonstrated poor performance and was prone to misclassification. We calculated the error rates within these regions by examining how often instances with these specific features were correctly classified versus misclassified.

\section{Result}

In this study, we employed the publicly available eICU dataset \cite{pollard2018eicu} to develop a predictive model for sepsis. The dataset comprises the vital signs of thousands of patients sampled at various rates. The vital signs considered in our experiments included systolic blood pressure, diastolic blood pressure, heart rate, respiration rate, oxygen saturation (SpO2), and gender of the patients. To standardize the sampling rates, all vital signs were resampled at a frequency of 5 minutes.
To build a classifier for this time-series data, we employed LightGBM. The time-series data was transformed into a format compatible with the LightGBM classifier \cite{ke2017lightgbm} by calculating rolling statistical properties such as mean, standard deviation, and lag values from previous timestamps. The model's parameters were optimized using Python library Optuna \cite{akiba2019optuna}. The model achieved a recall score of 0.9308 and 0.8125 on in-sample and out-of-sample splits.

 To gain a deeper understanding of the model's performance and identify areas where it exhibits suboptimal results, we applied the proposed method.
To visualize and communicate the regions of poor model performance, we present Fig. 1 a) demonstrates the error distribution over specific regions where the classifier exhibited suboptimal performance. Additionally, Fig. 1 b) provides a magnified view of the error distribution, offering a clearer resolution and facilitating a more detailed examination of the error rates within these identified regions. These figures serve as visual aids, aiding in the comprehension and interpretation of the model's performance shortcomings.

\begin{figure}[ht]
\centering
\includegraphics[width=1.0\textwidth]{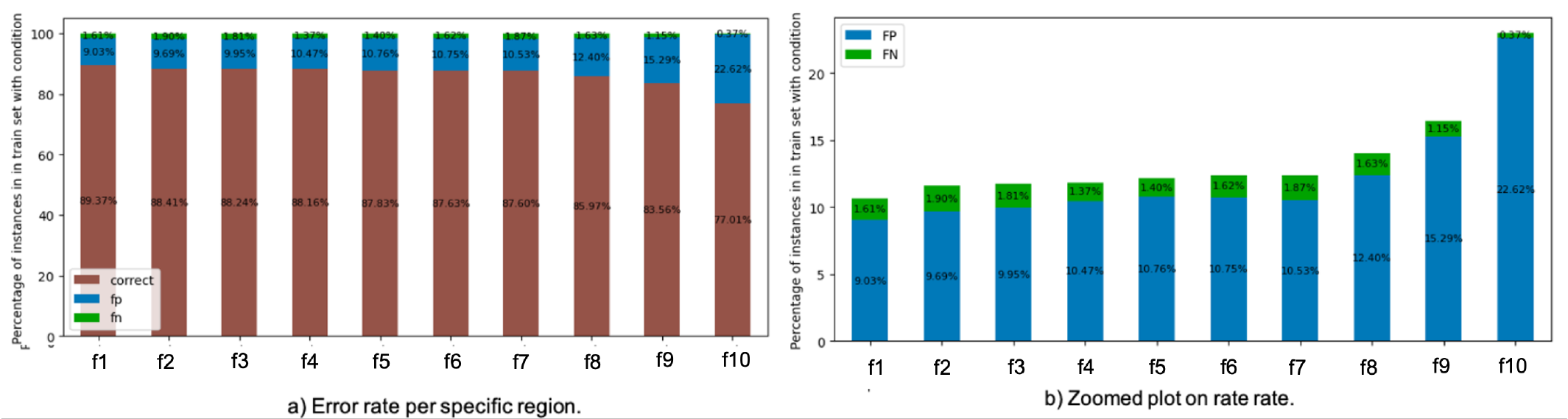}
\caption{Error Rate Plot for Feature Conditions (Regions) Impacting Poor Performance in Sepsis Detection Using eICU Dataset. }
\label{fig:error}
\end{figure}

In order to gain insights into the causes of misclassifications, we conducted a detailed analysis to determine which feature regions were most influential in contributing to these errors. Employing the LIME technique, we extracted the most significant features that consistently played a role in misclassification instances.

By identifying and examining these recurring features, we revealed specific regions where the classifier exhibited poor performance. Figure 1 visually illustrates the feature regions that meet this criterion, highlighting the factors associated with the model's suboptimal predictions.

% \begin{figure}[H]
% \subfigure[a]{\includegraphics[width=0.5\textwidth]{figures/output_XAI_1.png}}
% \subfigure[b]{\includegraphics[width=0.39\textwidth]{figures/output_XAI_2.png}}
% \caption{Distribution for each unique anchor}
% \label{fig:anchor_plots}
% \end{figure}

\section{Discussion}
In this study, we utilized LIME to identify regions where a black-box model exhibits poor performance. This approach allows us to investigate the error distribution across misclassification regions in both training and test data. The proposed method is model agnostic and can be utilized for any classifier. By analyzing the model's fit to the training data, we gain insights into its performance and identify areas where it inadequately represents the underlying patterns in the feature space. This assessment helps us understand the model's limitations in capturing the complexities of the training dataset.

When evaluating the model's generalization error on test data, we pinpoint specific regions within the feature space that contribute to erroneous predictions for unseen data. This knowledge is crucial for important decision-making situations, such as sepsis, where being aware of regions requiring caution is essential when relying on the classifier's predictions. By conducting this analysis, we obtain a comprehensive understanding of the model's limitations and areas of poor performance. This knowledge empowers healthcare professionals and decision-makers to make informed judgments, taking into account the regions in the feature space where the classifier's predictions may be less reliable.

\section{Conclusion}
Our study contributes to the understanding of machine learning models' performance by introducing a modified visualization approach that identifies regions of poor performance. By leveraging  LIME for the rule extraction method, we effectively pinpointed specific features responsible for misclassifications, allowing us to identify recurrent conditions associated with the classifier's suboptimal performance. The application of this methodology to the eICU dataset demonstrated its effectiveness in capturing regions where the classifier exhibits poor performance. These findings enhance interpretability and provide insights for decision-makers, enabling them to make informed choices regarding the deployment of machine learning models in critical domains such as sepsis detection.
In light of the study's insights, our future work aims to enhance the model's performance by  making specific modifications to the model architecture, feature engineering, and training strategies.

% \section{Acknowledgement}
% We would like to acknowledge the generous support of Happysona, an AI accelerator in Toronto, during the research of this project.
\bibliographystyle{splncs04}

\end{document}